\documentclass[11pt]{article}

\usepackage[final]{acl}

\usepackage{times}
\usepackage{latexsym}

\usepackage[T1]{fontenc}

\usepackage[utf8]{inputenc}

\usepackage{microtype}

\usepackage{inconsolata}

\usepackage{graphicx}

\usepackage{amsmath,amssymb}
\usepackage{booktabs}
\usepackage{comment}

\title{Objective-Specialized Residual Streams in Manifold-Constrained Hyper-Connections}

\author{
  Shantanu Agarwal \quad Ruchita V. Maitri \quad Zhiqi Huang \quad Vivek Datla \quad Alfy Samuel \\
  Capital One \\
  \small\texttt{\{shantanu.agarwal1, ruchitavishwanath.maitri, zhiqi.huang\}@capitalone.com} \\
  \small\texttt{\{vivek.datla, alfy.samuel\}@capitalone.com}
}

\begin{document}
\maketitle

\begin{abstract}
Manifold-constrained Hyper-Connections (mHC) extend a Transformer's residual stream into multiple parallel streams with learned cross-stream connections. Prior work trains all streams toward a shared objective. 
We instead study whether assigning distinct learning objectives to different streams can improve task performance and induce objective-specific specialization. 
Using complementary lexical (sparse) and semantic (dense) retrieval objectives as a test case, we find that objective-specialized mHC consistently outperforms a non-specialized mHC model across the full sparse--dense fusion range, for both two- and four-stream models and at two model scales. 
We then examine how specialization changes the model internally using probes of stream representations, residual mixing, objective-specific gradient flow, and read/write gates. 
The specialized and non-specialized models behave similarly through most of the network but diverge in the final layers: the specialized model develops increasingly objective-specific gradient flow, distinct read/write routing patterns, and reduced cross-stream mixing, and streams assigned to different objectives become increasingly dissimilar while those trained toward the same objective remain highly similar. 
Fixing the residual mixing to the identity leaves the specialization benefit intact, indicating that the benefit stems from objective-specific read/write routing rather than learned cross-stream mixing. 
These results show that assigning distinct objectives to mHC streams can improve multi-objective retrieval while inducing structured, depth-dependent specialization in the residual computation.
\end{abstract}

\section{Introduction}

Manifold-constrained Hyper-Connections (mHC)~\cite{xie2025mhcmanifoldconstrainedhyperconnections} replace a Transformer's single residual stream with multiple parallel streams~\cite{Zhu2024HyperConnections}, whose interactions are learned at each layer's attention and feed-forward modules. 
Existing applications of mHC jointly optimize all streams toward a shared objective~\cite{xie2025mhcmanifoldconstrainedhyperconnections,mhcpeft,mhc-gnn}, leaving open a basic question: \emph{what happens when different streams are assigned different objectives?}

Distinct objectives can benefit from shared representations and computation~\cite{Caruana1997MultitaskL,liu2019improvingmultitaskdeepneural}, while task-specific processing can be beneficial when objectives require different information~\cite{misra2016crossstitchnetworksmultitasklearning,sun2019learningsparsesharingarchitectures}. 
This motivates objective-specialized mHC, in which different streams are assigned distinct learning objectives while retaining the ability to interact through learned connections.

We investigate two questions. \emph{
RQ1: Can mHC streams specialize toward distinct learning objectives while improving joint task performance?
} (Section~\ref{sec:viability}) \emph{
RQ2: How does this specialization emerge across depth, as reflected in inter-stream representations, gradient flow, and read, write, and residual-mixing gates?
} (Section~\ref{sec:depth}).


As a concrete test case, we consider two complementary information-retrieval objectives: lexical and semantic matching~\cite{lin2023denserepresentationframeworklexical,lee-etal-2023-complementarity}. 
We train two- and four-stream mHC models, assigning each objective to a disjoint subset of streams, and evaluate at two model scales, ModernBERT-base and ModernBERT-large~\cite{warner2024smarterbetterfasterlonger}, to test whether the effect of specialization holds beyond a single model size. 
Across 13 BEIR benchmarks~\cite{thakur2021beirheterogenousbenchmarkzeroshot}, the pattern is consistent at both scales: objective specialization improves average nDCG@10 from $0.4310$ to $0.4486$ with two streams and from $0.4410$ to $0.4529$ with four streams at the base scale, and from $0.4574$ to $0.4667$ and from $0.4573$ to $0.4748$, respectively, at the larger scale. 
In addition, we demonstrate that the specialized models outperform their non-specialized counterparts across the full sparse--dense fusion range at both scales.

Beyond retrieval performance, we analyze how objective specialization emerges within the mHC stack using complementary measurements of stream routing, gradient flow, and inter-stream representations.
These diagnostics reveal a consistent depth-dependent pattern.
Through most of the Transformer, specialized and non-specialized models exhibit similar residual mixing and read/write routing, while gradients from the two objectives remain broadly distributed across streams.
In the final layers, however, the specialized models exhibit a distinct pattern absent in the non-specialized models: gradient flow becomes increasingly concentrated on the streams assigned to each objective, read/write routing becomes asymmetric, with the MLP increasingly reading from the sparse-objective streams and writing toward the dense-objective streams, and residual mixing moves toward the identity.
Additionally, the representation analysis provides a clear contrast: in the non-specialized model, all streams remain highly similar throughout the network, whereas in the specialized model, streams assigned to different objectives become increasingly dissimilar in the final layers.
These effects persist when increasing from two to four streams, suggesting that mHC supports shared computation at early depths followed by objective-specific processing near the output.
Finally, fixing the residual mixing to the identity leaves the specialization benefit intact, indicating that the benefit stems from the objective-specific read/write routing rather than from learned cross-stream mixing.

Together, these results show how objective specialization in mHC emerges across depth, connecting retrieval gains to changes in stream representations, gradient flow, and routing. To our knowledge, we are the first to assign distinct learning objectives to different mHC streams and characterize the resulting specialization across the Transformer stack.

\section{Related Work}

\subsection{Hyper-Connections}

Residual connections provide an identity path that allows representations to propagate across layers without transformation. 
Hyper-Connections (HC)~\cite{Zhu2024HyperConnections} generalize this paradigm by expanding the residual pathway into multiple parallel streams.
HC allows the multiple streams to interact through learned connectivity patterns, providing greater flexibility in how information propagates through depth. 
Manifold-Constrained Hyper-Connections (mHC)~\cite{xie2025mhcmanifoldconstrainedhyperconnections} constrain these learned connections to the Birkhoff polytope~\cite{Sinkhorn1967ConcerningNM}, ensuring that the residual mixing matrices remain doubly stochastic and thereby preserve the stability of the residual pathway while retaining learned cross-stream interactions.
Recent work has also begun to examine how the multiple residual streams introduced by HC and mHC are actually used after training ~\cite{alimaskina2026analyzingstreamcollapsehyperconnections,zhao2026doesmhcuseresidual}.
Our work builds on this multi-stream formulation but changes how the streams are trained: rather than jointly optimizing all streams toward the same objective, we assign different objectives to different streams and study the resulting organization of information flow across depth.

\subsection{Multi-Task and Multi-Objective Learning}

Multi-task learning aims to share computation across tasks while reducing task interference~\cite{Caruana1997MultitaskL,crawshaw2020multitasklearningdeepneural}. Prior work addresses this by learning task-specific representations or routing capacity, using approaches such as cross-stitch networks and mixture-of-experts~\cite{misra2016crossstitchnetworksmultitasklearning,10.1145/3219819.3220007,gupta2022sparselyactivatedmixtureofexpertsrobust}.
Our setting instead attaches distinct objectives directly to different mHC streams within the residual computation, allowing the streams to remain coupled while developing objective-specific behavior.

\subsection{Multi-Representation Retrieval}

Modern information retrieval increasingly combines representations that capture different notions of relevance. 
Sparse retrieval emphasizes lexical matching and exact or token-level correspondence, whereas dense retrieval captures semantic similarity through continuous representations~\cite{formal2021spladev2sparselexical,lin2023denserepresentationframeworklexical}. 
These signals are complementary: combining lexical and semantic matching can improve retrieval effectiveness and robustness over either representation alone~\cite{lin2023denserepresentationframeworklexical,lee-etal-2023-complementarity}.

Recent retrieval models demonstrate the value of combining complementary relevance signals within a single model. 
BGE-M3~\cite{chen-etal-2024-m3}, for example, supports dense, sparse, and multi-vector retrieval within a unified encoder, showing that multiple retrieval representations can be supported and efficiently combined. 
We instead ask whether distinct learning objectives can be assigned directly to different internal mHC streams, and how such specialization emerges across the Transformer stack. 
We use sparse and dense retrieval as complementary objectives to study this question.

\subsection{Retrofitting Pretrained Models with mHC}
Closely related to our setup is the use of mHC as a \emph{retrofit} onto an already-pretrained model. 
Oldenburg et al.~\citep{mhcpeft} wrap a pretrained model's residual stream with mHC for parameter-efficient fine-tuning. 
We likewise retrofit mHC onto a pretrained encoder, but differ in two respects: we fine-tune \emph{all} parameters (both the backbone and the mHC routing), and we assign distinct objectives to different streams rather than optimizing all streams toward a common objective.
\section{Task-Specialized Residual Streams in mHC}
\label{sec:mhc}

\subsection{Hyper connection}
\label{subsec:hc}
Whereas a standard residual network maintains a single hidden representation $x_l \in \mathbb{R}^d$ per token throughout the Transformer stack, Hyper-Connections (HC) \citep{Zhu2024HyperConnections} expand this hidden state into $n$ parallel streams.
We denote the hidden state of a token at layer $l$ as $X_l \in \mathbb{R}^{n \times d}$, where the $s$-th row vector $x_l^{(s)\top} \in \mathbb{R}^d$ represents the $s$-th stream for $s \in \{0, \dots, n-1\}$.
At the input layer ($l = 0$), the initial token embedding $x_0 \in \mathbb{R}^d$ is replicated across all $n$ streams, after which the streams propagate through the entire encoder stack.

At each Transformer sublayer $f_l(\cdot)$ (representing either Multi-Head Attention or a Feed-Forward Network), mHC \citep{xie2025mhcmanifoldconstrainedhyperconnections} govern information routing across streams via three components: a \emph{read} vector $p_l \in \mathbb{R}^n$ that determines how much each stream contributes to the sublayer's input, a \emph{write} vector $q_l \in \mathbb{R}^n$ that determines how much of the sublayer's output is written back into each stream, and a \emph{mixing} matrix $A_l \in \mathbb{R}^{n \times n}$ that governs how streams exchange information with one another along the residual path.

\paragraph{Aggregation (read).}
The read gate collapses the $n$ streams into a single vector using an element-wise weighted sum across streams.
This vector is the only input to the sublayer:
\begin{equation*}
u_l = \sum_{s=0}^{n-1} p_{l,s}\, x_l^{(s)} \in \mathbb{R}^{d},
\qquad
v_l = f_l(u_l) \in \mathbb{R}^{d}.
\end{equation*}

\paragraph{Distribution (write and mixing).}
The single sublayer output $v_l$ is split back into $n$ streams by the write gate, which assigns a per-stream scalar gain $q_{l,s}$ to $v_l$. In parallel, the mixing matrix recombines the incoming streams along the residual path, and the two are summed to form the next-layer streams:
\begin{equation*}
\begin{aligned}
x_{l+1}^{(s)}
&= \underbrace{\sum_{\tilde{s}=0}^{n-1} A_{l,s\tilde{s}}\,x_l^{(\tilde{s})}}_{\text{mixing of incoming streams}} \\
&\quad + \underbrace{q_{l,s}\,v_l}_{\text{write of sublayer output}}.
\end{aligned}
\label{eq:mhc-update}
\end{equation*}

\paragraph{Routing components.}
  The read, write, and mixing objects are not fixed~\citep{xie2025mhcmanifoldconstrainedhyperconnections}: they are produced from the current streams by learned linear projections with bias terms. Let
  $h_l = \mathrm{RMSNorm}\big([\,x_l^{(0)};\ldots;x_l^{(n-1)}\,]\big) \in \mathbb{R}^{nd}$ be the flattened, normalized streams. Then
  \begin{align*}
  p_l &= \sigma\!\big(\alpha^{\mathrm{pre}}\, W^{\mathrm{pre}} h_l + b^{\mathrm{pre}}\big), \\
  q_l &= 2\,\sigma\!\big(\alpha^{\mathrm{post}}\, W^{\mathrm{post}} h_l + b^{\mathrm{post}}\big), \\
  A_l &= \mathrm{Sinkhorn}\!\Big(\mathrm{reshape}\big(\alpha^{\mathrm{res}}\, W^{\mathrm{res}} h_l +
  b^{\mathrm{res}}\big)\Big),
  \end{align*} 
  where $p_l$, $q_l$, and $A_l$ are, respectively, the read, write, and mixing objects introduced above, $W^{\bullet}$ are learned projection weights, $b^{\bullet}$ are learned biases, and $\alpha^{\bullet}$ are learned scalar
  gains.
  The biases carry the static component of each gate, while the projections make the routing depend on the streams themselves.

\paragraph{Manifold constraint.} 
The defining feature of mHC is that the residual mixing matrix \(A_l\) is constrained to be doubly stochastic via Sinkhorn-Knopp normalization~\citep{Sinkhorn1967ConcerningNM}, which helps stabilize signal and gradient propagation across depth~\citep{xie2025mhcmanifoldconstrainedhyperconnections}.

\subsection{Encoder for Information Retrieval}
\label{sec:encoder}

\paragraph{Encoder.} We start from the publicly available, pretrained ModernBERT encoder~\citep{warner2024smarterbetterfasterlonger}, using both the base ($L=22$ layers, hidden size $d=768$)\footnote{\url{https://huggingface.co/answerdotai/ModernBERT-base}} and large ($L=28$ layers, $d=1024$)\footnote{\url{https://huggingface.co/answerdotai/ModernBERT-large}} variants, and fine-tune each for retrieval.
We retrofit mHC onto the pretrained encoder by wrapping every self-attention and MLP sublayer in an mHC block (Section~\ref{subsec:hc}) with $n \in \{2, 4\}$ streams.
The retrofit is function-preserving at initialization: for every layer $l$, the resulting stream $x_{l+1}^{(s)}$ is identical across all streams $s \in \{0,\dots,n-1\}$ and equal to the hidden state the pretrained ModernBERT model would produce at that layer (Appendix~\ref{appendix:mhc-init}). Fine-tuning therefore starts from the pretrained ModernBERT function, with any subsequent stream specialization learned during retrieval training.

\paragraph{Retrieval heads.}
Let $x_{L,t}^{(s)} \in \mathbb{R}^{d}$ denote the final-layer ($l=L$) contextual representation of token $t$ in stream $s$, where $s \in \{0,\ldots,n-1\}$; retrieval heads read only the final-layer streams.
Let $\mathcal{V}$ denote the set of content token positions, excluding special tokens.
Each retrieval head reads a designated subset of streams, $\mathcal{S}_{\mathrm{dn}}$ for the dense head and $\mathcal{S}_{\mathrm{sp}}$ for the sparse head.
The selected streams are mean-pooled at each token:
\begin{equation*}
\bar{x}_{L,t}^{\mathrm{dn}}
=
\frac{1}{|\mathcal{S}_{\mathrm{dn}}|}
\sum_{s\in\mathcal{S}_{\mathrm{dn}}} x_{L,t}^{(s)}
\quad
\bar{x}_{L,t}^{\mathrm{sp}}
=
\frac{1}{|\mathcal{S}_{\mathrm{sp}}|}
\sum_{s\in\mathcal{S}_{\mathrm{sp}}} x_{L,t}^{(s)}.
\label{eq:stream-avg}
\end{equation*}

The dense representation is obtained by masked mean pooling over content tokens followed by $L_2$ normalization:
\begin{equation*}
\tilde{e}^{\mathrm{dense}}
=
\frac{1}{|\mathcal{V}|}
\sum_{t \in \mathcal{V}}
\bar{x}_{L,t}^{\mathrm{dn}},
\quad
e^{\mathrm{dense}}
=
\frac{\tilde{e}^{\mathrm{dense}}}
{\left\|\tilde{e}^{\mathrm{dense}}\right\|_2}.
\label{eq:dense-head}
\end{equation*}

The sparse head produces a vocabulary-sized, non-negative lexical representation. 
A learned linear projection assigns a scalar importance to each token. 
The importance is rectified by ReLU and masked to zero on non-content tokens:
\begin{equation*}
w_t
=
\operatorname{ReLU}
\left(
u^{\top}\bar{x}_{L,t}^{\mathrm{sp}} + b
\right),
\label{eq:sparse-token-weight}
\end{equation*}
where $u \in \mathbb{R}^{d}$ and $b \in \mathbb{R}$ are learned parameters and $w_t = 0$ for $t \notin \mathcal{V}$.
Let $\mathrm{id}(t) \in \{1,\ldots,|\mathcal{W}|\}$ denote the vocabulary id of token $t$, where $\mathcal{W}$ is the vocabulary. 
The token weights are then max-pooled according to vocabulary id:
\begin{equation*}
e_v^{\mathrm{sparse}}
=
\max_{\substack{t \in \mathcal{V}\\ \mathrm{id}(t)=v}}
w_t,
\qquad
v \in \{1,\ldots,|\mathcal{W}|\},
\label{eq:sparse-head}
\end{equation*}
 If vocabulary id $v$ does not occur among the content tokens, we set $e_v^{\mathrm{sparse}}=0$. 
 Thus, the sparse representation contains non-zero weights only for terms occurring in the input, following the lexical term-weighting formulation of uniCOIL~\citep{lin2021briefnotesdeepimpactcoil}, BGE-M3~\citep{chen-etal-2024-m3}.

\paragraph{Scoring.}
Let $q$ and $d$ denote a query and a candidate document, respectively. 
Both are encoded with the \emph{same} encoder, each independently producing representations $e_q^{\mathrm{sparse}}, e_q^{\mathrm{dense}}$ and $e_d^{\mathrm{sparse}}, e_d^{\mathrm{dense}}$ via the retrieval heads above. Retrieval scores are computed as inner products of matching representation types:
\begin{equation}
s^{o}(q,d) = \langle e_q^o, e_d^o \rangle,
\qquad o \in \{\mathrm{sparse},\mathrm{dense}\}.
\label{eq:retrieval-scores}
\end{equation}

\paragraph{Stream specialization.}

\begin{figure*}[t]
\centering
\includegraphics[width=1.0\textwidth]{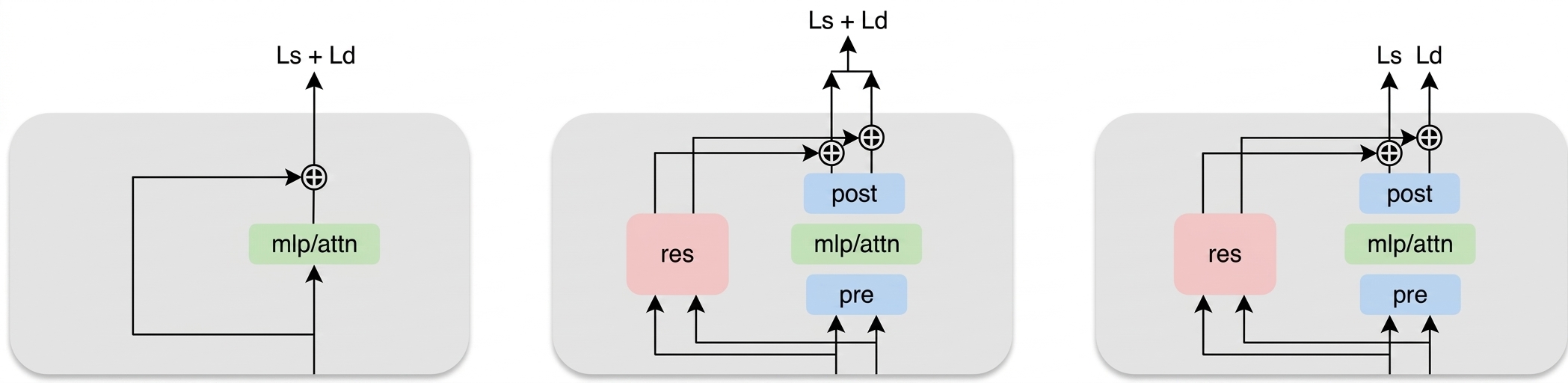}
\caption{
Schematic comparison of the three encoder configurations studied in this work. In all three, streams propagate through the full encoder stack and the retrieval heads read only the \emph{final-layer} ($l=L$) stream representations. 
\textbf{(left) Single-stream}: the standard ModernBERT encoder, with a single residual stream read by both the sparse and dense retrieval heads. 
\textbf{(middle) Non-specialized mHC}: the residual stream is expanded into $n$ parallel streams with learned cross-stream routing. Both retrieval heads operate on the mean-pooled representation obtained from all $n$ streams, so the sparse and dense objectives jointly optimize the same stream representation.
\textbf{(right) Objective-specialized mHC}: the sparse and dense heads instead read disjoint sets of final-layer streams, so each objective selectively trains its assigned stream(s).
For clarity the schematic depicts $n=2$ streams; we study both $n=2$ and $n=4$ in our experiments.
}
\label{fig:mhc}
\end{figure*}

We consider three configurations that differ only in which streams are read by the retrieval heads. In the \textbf{objective-specialized} configuration ($n \in \{2,4\}$), $\mathcal{S}_{\mathrm{sp}}=\{0,\ldots,n/2-1\}$ and $\mathcal{S}_{\mathrm{dn}}=\{n/2,\ldots,n-1\}$: the sparse and dense retrieval heads read disjoint stream groups, allocating the first half of the streams exclusively to the sparse objective and the second half to the dense objective. 
This decouples task readouts at the output layer while mHC routing maintains inter-stream coupling throughout the encoder stack. 
In the \textbf{non-specialized} configuration ($n \in \{2,4\}$), $\mathcal{S}_{\mathrm{sp}}=\mathcal{S}_{\mathrm{dn}}=\{0,\ldots,n-1\}$: both retrieval heads read the mean-pooled representation across all streams, forcing both objectives to jointly optimize a shared representation space. 
Finally, in the \textbf{single-stream} configuration ($n=1$), $\mathcal{S}_{\mathrm{sp}}=\mathcal{S}_{\mathrm{dn}}=\{0\}$: with a single stream, there is no cross-stream routing and the encoder is exactly the original ModernBERT model, providing the dedicated single-objective reference points.

\subsection{Training Setup}
\label{sec:training}

\paragraph{Training data.}
We fine-tune all models on the Natural Questions (NQ)~\cite{kwiatkowski-etal-2019-natural} dataset. 
We use the first $90$k training pairs for fine-tuning and reserve the remaining pairs as a validation set for checkpoint selection\footnote{\href{https://huggingface.co/datasets/malteos/natural-questions}{Natural Questions} has $\sim 100\mathrm{K}$ queries.}. 
Queries and passages are truncated to $32$ and $256$ tokens, respectively\footnote{The median query and document lengths are approximately 10 and 100 whitespace-delimited words, respectively.}.

\paragraph{Dense and sparse objectives.}
We optimize both sparse and dense representations using a contrastive retrieval loss~\citep{oord2019representationlearningcontrastivepredictive} with in-batch negatives. Let $s^{o}(q,d)$ denote the relevance score between query $q$ and document $d$ for retrieval type $t \in {\mathrm{dense},\mathrm{sparse}}$. For a batch of $B$ query--positive pairs, the per-objective loss is
\begin{equation}
\mathcal{L}_{o}
= -\frac{1}{B}\sum_{i=1}^{B}
  \log \frac{\exp\!\big(s^{o}(q_i, p^{+}_i)/\tau_t\big)}
            {\sum_{d \in \mathcal{C}} \exp\!\big(s^{o}(q_i, d)/\tau_t\big)},
\end{equation}
where $\tau_{\mathrm{dense}}$ and $\tau_{\mathrm{sparse}}$ are fixed temperature hyperparameters and $\mathcal{C}$ is the batch document collection. 
Thus, for each query $q_i$, its paired passage $p_i^{+}$ is treated as the positive, while the remaining $B-1$ passages in the batch serve as in-batch negatives.

\paragraph{Combined training loss.}
The multi-objective models are trained on the sum of the two objectives,
\begin{equation}
\mathcal{L} = \mathcal{L}_{\mathrm{dense}} + \lambda_{\mathrm{sparse}}\,\mathcal{L}_{\mathrm{sparse}},
\label{eq:combined}
\end{equation}
where $\lambda_{\mathrm{sparse}} \ge 0$ sets the weight of the sparse objective relative to the dense one. 
Single-objective baselines use only the corresponding term. In the specialized model, $\mathcal{L}_{\mathrm{dense}}$ and $\mathcal{L}_{\mathrm{sparse}}$ are computed from disjoint streams (Section~\ref{sec:encoder}), so each term supervises its own set of streams.

\paragraph{Training parameters.}
All model variants are fine-tuned using identical optimization setups: the Adam optimizer~\citep{kingma2017adammethodstochasticoptimization} with no weight decay, a linear warmup-and-decay learning-rate schedule with a $5\%$ warmup phase, gradient clipping at a maximum norm of $1.0$, and \texttt{bfloat16} mixed precision. 
Each model is trained for an epoch on two NVIDIA H100 GPUs. 
We evaluate checkpoints every $250$ steps on the held-out NQ validation set and select the checkpoint with the highest mean triplet accuracy across the two retrieval heads. 
For each objective $o \in \{\mathrm{dense},\mathrm{sparse}\}$, triplet accuracy is the fraction of validation triplets satisfying $s^{o}(q,d^+) > s^{o}(q,d^-)$.
To ensure a fair comparison, all model families are optimized over the same hyperparameter grid: learning rate $\eta \in \{5\times10^{-5}, 8\times10^{-5}, 1\times10^{-4}\}$, dense loss temperatures $\tau_{\mathrm{dense}} \in \{0.01,0.05\}$, sparse loss temperatures $\tau_{\mathrm{sparse}} \in \{0.01,0.05\}$, sparse loss weight $\lambda_{\mathrm{sparse}} \in \{0.1\}$, and batch size $B \in \{64,128\}$.

\paragraph{Inference and Hybrid Fusion.}
At inference, the dense and sparse heads act as independent retrievers: each scores candidates via $s^{o}(q,d)$ (Eq.~\ref{eq:retrieval-scores}) within its own feature space and returns its top-$k$ candidates. 
We fuse the two rankings with a score-based combination over the union of candidates~\citep{chen-etal-2024-m3}:
\begin{equation}
s_{\alpha}(q,d)
=
\alpha\, s^{\mathrm{dense}}(q,d)
+
(1-\alpha)\, s^{\mathrm{sparse}}(q,d),
\label{eq:fusion}
\end{equation}
where $\alpha \in [0,1]$ and $s^{o}(q,d)=0$ if $d$ is outside head $o$'s top-$k$ candidates\footnote{We set $k=1000$ in all experiments.}. 
The endpoints recover single-head retrieval ($\alpha=0$: pure sparse; $\alpha=1$: pure dense).
We evaluate performance across the full continuum $\alpha \in [0,1]$ (Section~\ref{sec:viability}).

\section{Objective-Specialized Residual Streams}
  \label{sec:viability}

\begin{table*}[t]
    \centering
    \small
    \resizebox{\textwidth}{!}{%
    \begin{tabular}{lcccccccccccccc}
    \toprule
    Model & ArguAna & C-FEVER & DBPedia & FEVER & FiQA & HotpotQA & NFC & NQ &
    Quora & SciDocs & SciF & TREC-C & Touch\'e & Avg \\
    \midrule
    \multicolumn{15}{l}{$n=1$} \\
    Sparse-only & 0.3943 & 0.1052 & 0.1484 & 0.6384 & 0.1953 & \textbf{\underline{0.5895}} &
    0.2219 & 0.1600 & 0.6722 & 0.1119 & 0.6492 & 0.5381 & 0.2372 & 0.3585 \\
    Dense-only & \underline{0.3974} & 0.2116 & \underline{0.2702} & 0.5049 & 0.2373 & 0.3797 & 0.2460 & \underline{0.4569} & \underline{0.8479} & \underline{0.1328} & 0.5513 & 0.6784 & 0.2201 & 0.3950 \\
    Sparse+Dense & 0.3833 & \underline{0.2217} & 0.2552 & \underline{0.6525} & \underline{0.2649} & 0.5678 &
    \underline{0.2611} & 0.4110 & 0.8274 & 0.1310 & \underline{0.6707} & \underline{0.6990} & \underline{0.2643} & \underline{0.4315} \\
    \midrule
    \multicolumn{15}{l}{$n=2$} \\
    mHC no-spec & 0.3827 & 0.2132 & 0.2579 & \underline{0.6529} & 0.2718 & \underline{0.5696} &
    0.2708 & 0.4079 & 0.8229 & 0.1346 & \underline{0.6709} & 0.6843 & \underline{0.2634} & 0.4310 \\
    mHC spec & \underline{0.4278} & \underline{0.2394} & \underline{0.2919} & 0.6294 &
    \textbf{\underline{0.2885}} & 0.5226 & \underline{0.2767} & \underline{0.4681} &
    \underline{0.8483} & \textbf{\underline{0.1535}} & 0.6672 & \underline{0.7585} &
    0.2598 & \underline{0.4486} \\
    \midrule
    \multicolumn{15}{l}{$n=4$} \\
    mHC no-spec & 0.4068 & 0.2304 & 0.2584 & \textbf{\underline{0.6671}} & 0.2752 &
    \underline{0.5716} & 0.2694 & 0.4154 & 0.8301 & 0.1366 & \textbf{\underline{0.6751}} &
    0.7280 & \textbf{\underline{0.2688}} & 0.4410 \\
    mHC spec & \textbf{\underline{0.4287}} & \textbf{\underline{0.2419}} & \textbf{\underline{0.2991}} & 0.6385 &
    \underline{0.2872} & 0.5226 & \textbf{\underline{0.2848}} & \textbf{\underline{0.4757}} &
    \textbf{\underline{0.8516}} & \underline{0.1520} & 0.6689 & \textbf{\underline{0.7692}} & 0.2676 &
    \textbf{\underline{0.4529}} \\
    \bottomrule
    \end{tabular}%
    }
    \caption{
    BEIR nDCG@10 across all 13 datasets using ModernBERT-base. 
    Sparse-only and Dense-only are evaluated in their native retrieval mode; Sparse+Dense and the mHC models are hybrid retrievers fused at $\alpha=0.9$, the shared empirical optimum across all multi-objective models. 
    \underline{Underline} marks the best score within each stream count $n$; \textbf{bold} marks the best score for that dataset across all configurations.
    The advantage of objective specialization is not confined to the aggregate score; it extends to a majority of individual datasets and holds at both $n=2$ and $n=4$.
    Excluding the in-domain NQ dataset, \emph{mHC+spec} still improves over \emph{mHC no-spec} on the remaining 12 datasets ($0.4329\!\to\!0.4470$ at $n{=}2$; $0.4431\!\to\!0.4510$ at $n{=}4$). 
    Results reported here are mean over two seeds. Per seed variation is provided in Appendix \ref{app:seeds}.
   Results with ModernBERT-large (Table~\ref{tab:beir-large}) show the same trend: mHC+spec improves over mHC no-spec at both $n=2$ and $n=4$, indicating that the benefit of objective specialization is robust to model scale rather than specific to a single model size.}
    \label{tab:beir-categories}
\end{table*} 

To address RQ1, we ask whether mHC residual streams can specialize to distinct learning objectives while improving multi-objective retrieval. 
We compare objective-specialized mHC (\emph{mHC+spec}) against three baselines: (1) \textbf{dedicated single-objective models}, which provide single-task reference points; (2) a \textbf{standard non-mHC joint model} (\emph{Sparse+Dense}), which combines the two objectives within a conventional single residual stream; and (3) a \textbf{non-specialized mHC model} (\emph{mHC no-spec}), in which both retrieval heads read the pooled mHC streams. 
The third comparison is the key ablation: \emph{mHC no-spec} and \emph{mHC+spec} use the same architecture, training data, and optimization budget, differing only in whether the objectives are assigned to distinct streams. We evaluate both $n=2$ and $n=4$ streams. 
We evaluate our system on 13 BEIR datasets \cite{thakur2021beirheterogenousbenchmarkzeroshot} and report results averaged over two training seeds.

\subsection{Specialization improves retrieval across fusion weights}
\label{sec:alpha}

\begin{figure}[h!]
    \centering
    \includegraphics[width=\columnwidth]{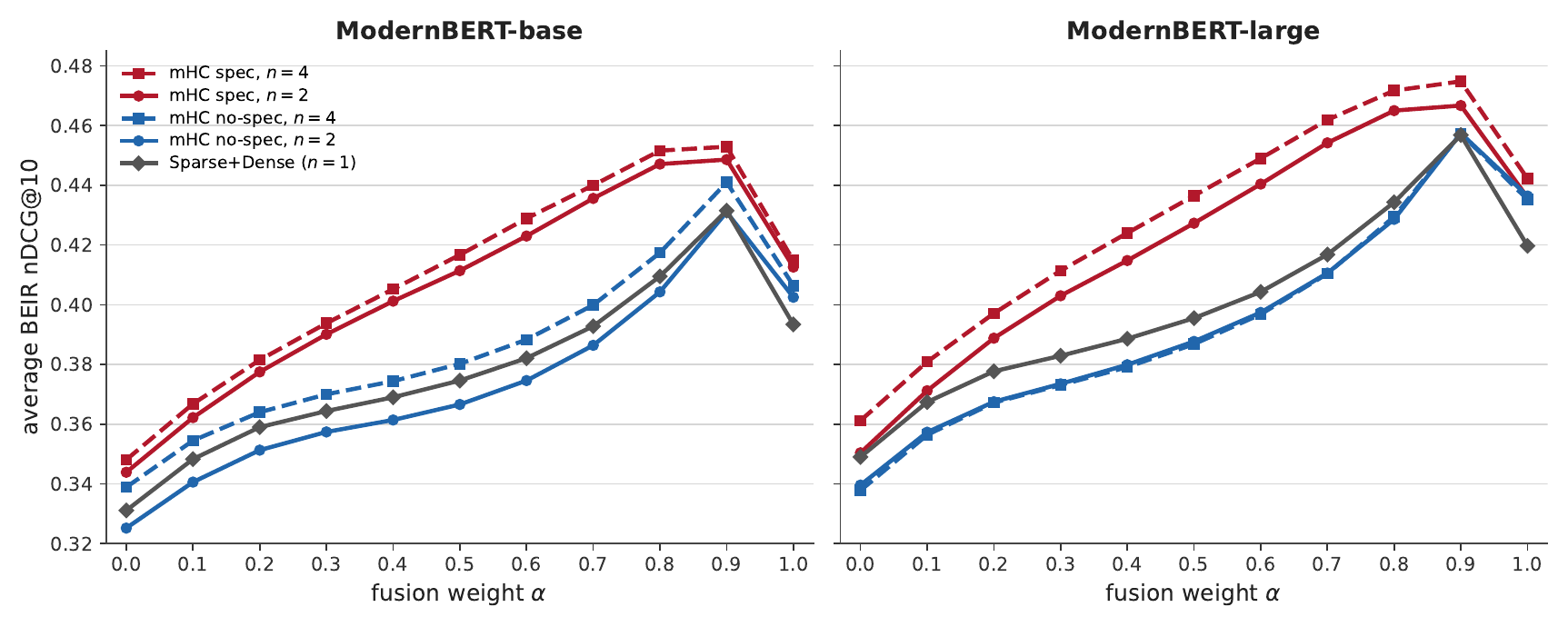}
    \caption{
    Average BEIR nDCG@10 as a function of the fusion weight $\alpha$, for ModernBERT-base (left) and ModernBERT-large (right); both mean over two seeds. 
    $\alpha=0$ and $\alpha=1$ correspond to pure sparse and pure dense retrieval, respectively.
    Color encodes objective specialization (red: specialized, blue: non-specialized); Sparse+Dense ($n{=}1$, grey) has no stream routing. Specialized models (red) dominate at every $\alpha$ for both stream counts and at both base/large model scales.
    }
    \label{fig:beir-alpha-sweep}
\end{figure}

Because the multi-objective models produce both sparse and dense representations, we evaluate retrieval across the full sparse--dense fusion range using the score fusion defined in Eq.~\ref{eq:fusion}, from pure sparse ($\alpha=0$) to pure dense ($\alpha=1$). Figure~\ref{fig:beir-alpha-sweep} reports average BEIR nDCG@10 across this sweep.

The results show three findings. First, at the base scale, four-stream mHC models outperform the conventional \emph{Sparse+Dense} model, while the two-stream models are comparable. This comparison is secondary, since mHC introduces additional stream-specific parameters and routing capacity\footnote{mHC introduces a modest parameter overhead relative to the corresponding single-stream model: approximately $0.41\%$ for $n=2$ and $2.25\%$ for $n=4$.}. Second, and more importantly, \emph{mHC+spec} consistently outperforms \emph{mHC no-spec} at every fusion weight for both $n=2$ and $n=4$; Figure~\ref{fig:beir-alpha-sweep} shows that this also holds at a larger model scale (ModernBERT-large). At $\alpha=0.9$ and the base scale, specialization improves average nDCG@10 from $0.4310$ to $0.4486$ with two streams and from $0.4410$ to $0.4529$ with four streams (ModernBERT-large: $0.4574 \to 0.4667$ and $0.4573 \to 0.4748$, respectively). Thus, the improvement is not explained simply by the additional capacity of mHC, but by how that capacity is organized across objectives. Third, the specialization benefit persists across the full sparse--dense fusion range rather than depending on a particular operating point, at both model scales.

All five multi-objective models achieve their best average score at $\alpha=0.9$ at both scales, which we use for the per-dataset analysis below. At the base scale, increasing the number of streams also improves both specialized and non-specialized mHC, indicating complementary benefits from additional stream capacity and objective specialization.

\subsection{Per-dataset multi-objective retrieval}
\label{sec:mo-retrieval}

Table~\ref{tab:beir-categories} reports seed-averaged nDCG@10 on all $13$ BEIR datasets at $\alpha=0.9$ using ModernBERT-base; the corresponding per-dataset results for ModernBERT-large are reported in Appendix~\ref{app:large}.
The advantage of objective specialization is not confined to the aggregate score: mHC+spec improves over mHC no-spec on $9$ of the $13$ datasets.
The improvements are not universal, and the exceptions are the same across stream counts: on FEVER, HotpotQA, SciFact, and Webis-Touch\'e2020.

\section{How specialization reshapes the streams}
\label{sec:depth}

\begin{figure*}[ht]
      \centering
      \includegraphics[
      width=\textwidth,
      height=1.0\textheight,
      keepaspectratio,
      ]{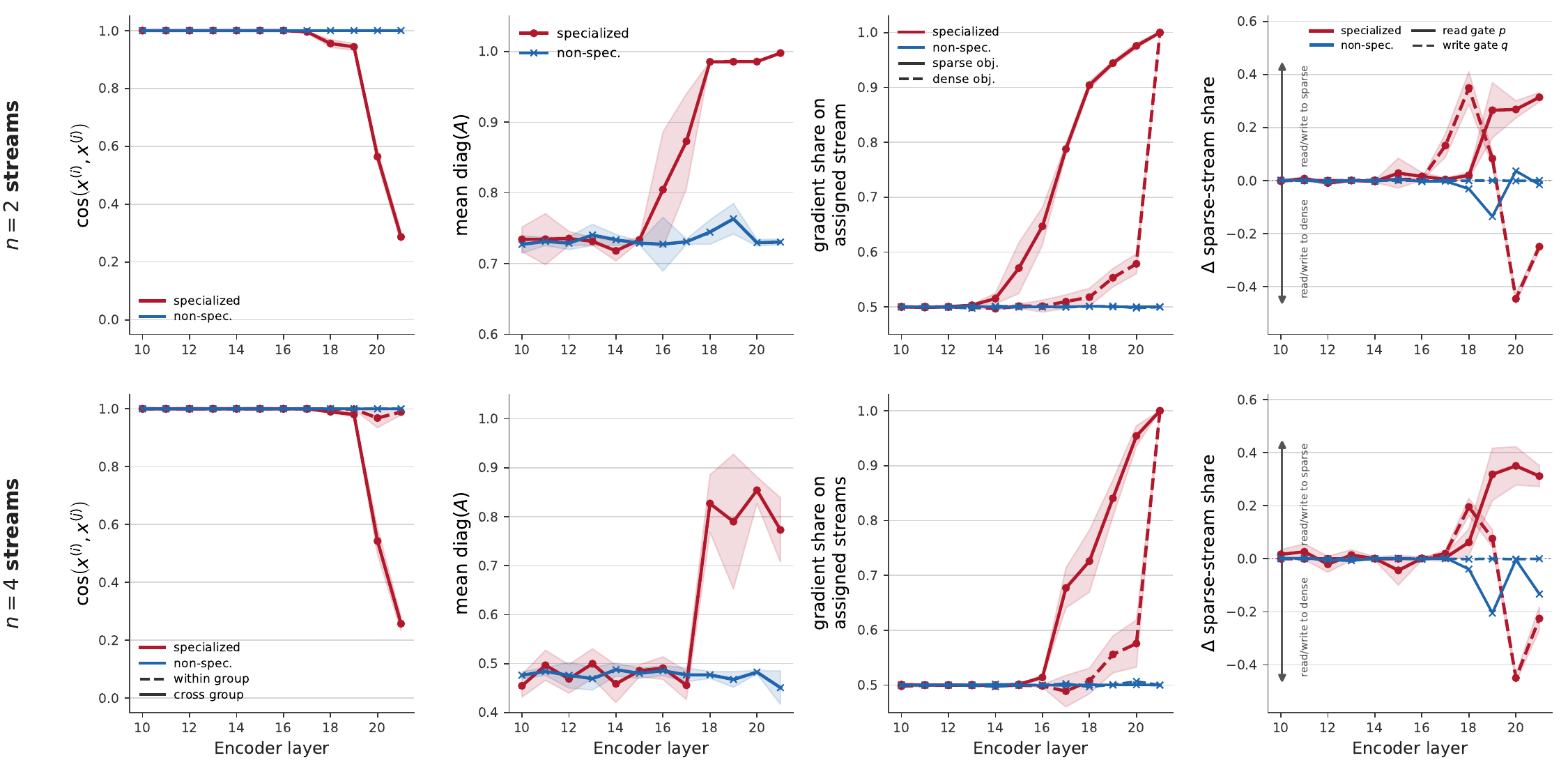}
      \caption{
      Four depth-resolved specialization probes at the final checkpoint, for $n{=}2$ (top) and $n{=}4$ (bottom) streams (lines are the mean and
  shaded bands $\pm 1$ standard deviation over 4 seeds; flat layers $0$--$9$ omitted). 
      From left: stream cosine similarity, residual-mixing identity retention $\mathrm{diag}(A)$, per-objective gradient share on the assigned stream(s), and read/write gate shift. 
      Red: specialized; blue: non-specialized control. 
      Solid vs.\ dashed distinguishes the two objectives / gates / stream-pair types (legends).
      Metrics are defined and discussed in Section~\ref{sec:depth}.
      }
      \label{fig:dynamics-grid}
  \end{figure*}
Since the specialized and non-specialized models have the same architecture and capacity, their performance difference raises a more specific question: \emph{how does assigning different objectives to the streams change the internal computation?} 
We therefore ask whether, where, and how the two models diverge \emph{internally} (\textbf{RQ2}). 
The models share the same architecture, training data, and optimization budget and differ only in whether or not the retrieval heads read disjoint stream groups, allowing us to attribute systematic internal differences to objective specialization.

We examine four complementary signals across encoder depth $\ell$ at the final checkpoint: (1) stream cosine similarity, which measures whether the stream representations diverge; (2) residual mixing, which measures how strongly information is exchanged between streams; (3) objective-specific gradient flow, which measures whether each objective preferentially trains its assigned stream; and (4) read and write gate shifts, which reveal how routing in MLP and attention submodules change to support this specialization.
All curves are means over four seeds; the lower layers ($\ell<10$) are flat across probes and are omitted. 
Figure~\ref{fig:dynamics-grid} summarizes the four depth-resolved probes, measured at the MLP sublayers. The corresponding results for the attention sublayers for residual mixing $\mathrm{diag}(A)$ and read/write gates are provided in Appendix~\ref{app:probe-attn}.

\paragraph{Representational divergence.}
We first ask whether objective specialization produces different stream representations. 
At each layer, we compute the cosine similarity between token-averaged stream states. 
The non-specialized models remain highly similar across streams, whereas the specialized models increasingly separate streams assigned to different objectives in the final layers. 
With two streams, cosine similarity falls to $\approx 0.3$ at layer~21; with four streams, streams assigned to the same objective remain nearly identical ($\cos\!\approx\!0.99$), while cross-objective pairs diverge ($\cos\!\approx\!0.26$) (Figure~\ref{fig:dynamics-grid}). 
Thus, specialization organizes the streams into objective-specific groups rather than simply replicating the same representation across all streams.
\footnote{We use cosine similarity as the primary measure of inter-stream similarity. We also evaluated linear CKA, but it remained close to $1$ across the network and did not show the depth-dependent separation observed with cosine similarity.}

\paragraph{Reduced residual mixing.}
We next ask whether the streams remain strongly coupled through the residual mixing matrix $A$ as their representations diverge. 
We measure the mean diagonal $\frac{1}{n}\sum_i A_\ell[i,i]$ of the doubly stochastic matrix $A_\ell$, where values near $1$ indicate that $A_\ell$ approaches the identity and cross-stream mixing is reduced. 
In the specialized model, $\mathrm{diag}(A_\ell)$ increases sharply in the final layers, approaching $1.0$ for $n=2$ and reaching $\approx 0.8$ for $n=4$; the non-specialized model remains near its initialization value ($0.73$ for $n=2$, $0.48$ for $n=4$) throughout.
Thus, representational divergence is accompanied by a marked reduction in cross-stream mixing in the specialized model.
Section~\ref{sec:identity-res} tests whether this reduction is needed for the specialization benefit.

\paragraph{Objective-specific gradient flow.}
We next ask whether representational divergence is accompanied by objective-specific training signals. For each objective $o\in\{\mathrm{sparse},\mathrm{dense}\}$, we back-propagate its loss separately and measure the gradient magnitude it delivers to each stream, $k$, at layer $\ell$:
\begin{equation*}
g^{o}_\ell[k]
=
\frac{\left\lVert \partial\mathcal{L}_o/\partial x^{(k)}_\ell\right\rVert}
{\sum_j\left\lVert \partial\mathcal{L}_o/\partial x^{(j)}_\ell\right\rVert},
\qquad
\sum_k g^{o}_\ell[k]=1.
\end{equation*}
We report the \emph{assigned-stream share}, $\sum_{k\in\mathcal{S}_o} g^{o}_\ell[k]$, where $\mathcal{S}_o$ is the set of streams assigned to objective $o$ (for $n{=}2$, $\mathcal{S}_{\mathrm{sparse}}{=}\{0\}$ and $\mathcal{S}_{\mathrm{dense}}{=}\{1\}$; for $n{=}4$, $\{0,1\}$ and $\{2,3\}$). 
A value near $|\mathcal{S}_o|/n = 0.5$ indicates that the objective trains all streams equally, whereas a value near $1$ indicates concentration on its assigned stream. 
In the specialized model, both objectives increasingly concentrate their gradients on their assigned streams in the upper layers. 
The sparse objective localizes earlier and more sharply, while the dense objective remains more distributed until the final layers. 
For the non-specialized control no retrieval head reads a designated subset of streams, so no stream is intrinsically tied to an objective. 
We nonetheless evaluate the probe using the \emph{same} index sets $\mathcal{S}_o$ that specialization would impose, treating the partition as a fixed measurement convention rather than a property of the model. 
Because nothing links either objective to those streams, each objective's gradient spreads evenly across all streams and the share stays near $0.5$ throughout depth. 
This flat profile is the appropriate null reference: it is exactly what the probe returns when the stream partition carries no objective-specific information, so that any departure from it in the specialized model reflects specialization rather than a property of the metric.

\paragraph{Read/write gate specialization.}
Finally, we examine how the MLP routing reorganizes to support this objective-specific processing. 
Recall that at each layer the read gate emits a vector of per-stream read weights $p_\ell\in\mathbb{R}^{n}$ and the write gate a vector of per-stream write weights $q_\ell\in\mathbb{R}^{n}$ (Section~\ref{sec:mhc}). 
For a gate $g_{\ell}\in\{p_\ell, q_\ell\}$ we define its \emph{sparse-stream share} as the fraction of its total mass placed on the stream(s) assigned to the sparse objective,
\begin{equation*}
s(g_{\ell}) \;=\; \frac{\sum_{k\in\mathrm{sparse}} g_{\ell}[k]}{\sum_{j\in\{\mathrm{sparse}, \mathrm{dense}\}} g_{\ell}[j]},
\end{equation*}
which for $n=2$ reduces to the weight on the single sparse-assigned stream. 
Because our initialization places a small amount of random mass on each stream, we report the \emph{change} in this share from initialization to the final checkpoint, $\Delta s(g_{\ell}) = s(g_{\ell})\big|_{T} - s(g_{\ell})\big|_{0}$, where $T$ denotes the final training step.
This is the quantity plotted in Figure~\ref{fig:dynamics-grid}. 
A positive $\Delta s$ means the gate (read or write) moves mass toward the sparse-assigned stream, a negative value toward the dense-assigned stream.

In the specialized models, this pattern holds for both $n=2$ and $n=4$: the read gate increases its sparse-stream share ($\Delta s(p_{L})\approx+0.3$) while the write gate decreases it ($\Delta s(q_{L})\approx-0.25$). 
Since both gates start near an even split, the MLP ends up reading predominantly from the sparse-assigned streams and writing predominantly to the dense-assigned streams. 
The non-specialized controls show essentially no shift in either gate ($\Delta s\approx0$).

Taken together, the four probes reveal a consistent depth-dependent pattern. The specialized and non-specialized models remain similar through most of the network, but diverge in the final layers: streams assigned to different objectives become more distinct, objective-specific gradients concentrate on their assigned streams, the gates develop a directional read/write pattern (preferentially reading from sparse-assigned streams and writing to dense-assigned streams), and residual mixing decreases.
The results therefore connect the downstream benefit of objective specialization to a concrete change in the internal computation of mHC rather than to a difference in capacity alone.

\subsection{Read/write routing, not mixing, drives the specialization benefit}
\label{sec:identity-res}

\begin{table}[t]
\centering
\small
\begin{tabular}{llc}
\toprule
Model & Mixing $A$ & nDCG@10 ($\alpha{=}0.9$) \\
\midrule
mHC no-spec & learned  & $0.4410 \pm 0.0023$ \\
mHC spec    & learned  & $0.4529 \pm 0.0019$ \\
\midrule
mHC no-spec & identity & $0.4415 \pm 0.0025$ \\
mHC spec    & identity & $0.4546 \pm 0.0001$ \\
\bottomrule
\end{tabular}
\caption{Fixing the residual-mixing matrix to the identity ($A_\ell{=}I$ at every layer). ModernBERT-base, $n{=}4$, average BEIR nDCG@10 over 13 datasets at $\alpha{=}0.9$; mean $\pm$ standard deviation over two seeds.}
\label{tab:identity-res}
\end{table}

Specialization drives the residual-mixing matrix toward the identity in the final layers. 
This suggests a hypothesis: the specialization benefit stems from the objective-specific read/write routing rather than from learned cross-stream mixing. 
To test it, we fix $A_\ell = I$ at every layer, so that each stream is carried forward unmixed and streams interact only through the read and write gates, and retrain both the specialized and non-specialized models with $n=4$ (Table~\ref{tab:identity-res}).
Fixing the mixing to the identity changes average nDCG@10 by less than the variation across seeds, and this holds with or without specialization: mHC no-spec scores $0.4410$ with learned and $0.4415$ with identity mixing, and mHC+spec scores $0.4529$ and $0.4546$, respectively.
Crucially, the specialization gain is preserved ($+0.0119$ with learned mixing, $+0.0131$ with identity mixing), well beyond the seed-to-seed standard deviation.
These results support the hypothesis: learned cross-stream mixing is not required for the specialization benefit, which instead stems from the objective-specific read/write routing.
\section{Conclusion}

We studied whether mHC residual streams can be assigned distinct learning objectives, using sparse and dense retrieval as a testbed.
Specialization consistently improves retrieval over non-specialized mHC models of identical capacity, training data, and optimization budget, across the full sparse--dense fusion range and for both $n{=}2$ and $n{=}4$ streams, and this holds at both ModernBERT-base and ModernBERT-large scales.
The specialized models also improve over the conventional single-stream joint retriever, with the $n{=}4$ model achieving the best BEIR score at both scales.

Our analyses further show that specialization emerges primarily in the final layers.
There, stream representations diverge, objective-specific gradients concentrate on assigned streams, the MLP gates develop a read-from-sparse / write-to-dense pattern, and residual mixing moves toward the identity.
At $n{=}4$, streams assigned to the same objective remain similar, forming two coherent objective groups.
Fixing the residual mixing to the identity leaves the specialization benefit intact, indicating that it stems from the objective-specific read/write routing rather than from learned cross-stream mixing.
Together, these results show that assigning distinct objectives to mHC streams can both improve downstream performance and induce structured, depth-localized specialization within the streams.
\section*{Limitations}

\paragraph{Objective diversity.}
We study a single pair of objectives---lexical (sparse) and semantic (dense) retrieval---chosen because they are related but distinct. We do not systematically vary the degree of objective divergence, and therefore cannot determine how specialization changes as objectives become more similar, complementary, or conflicting.

\paragraph{Task scope.} Our experiments cover only two scales (ModernBERT-base and ModernBERT-base) with $n \in \{2,4\}$ streams, and encoder-based retrieval. We therefore do not establish whether the observed performance gains or late-layer specialization patterns extend to other model scales, other numbers of streams or objectives, or decoder-based and generative tasks.

\paragraph{Equal allocation of stream capacity across objectives.}
In our specialized mHC models, we allocated an equal number of streams to the sparse and dense objectives. 
We leave the effect of asymmetric stream allocation on downstream performance to future work.


\bibliography{custom}

\appendix
\section{mHC routing parameter initialization}
\label{appendix:mhc-init}

For initialization choices, we follow prior work on retrofitting pretrained Transformers with mHC~\citep{mhcpeft}.
Each mHC block is initialized so that, at the start of fine-tuning, its routing is \emph{static} (independent of the input token's stream representations) and the block behaves like an ordinary residual sublayer (Section~\ref{subsec:hc}). 
The projection weights for all three routing components are zero-initialized ($W^{\mathrm{pre}}, W^{\mathrm{post}}, W^{\mathrm{res}} = 0$), so each gate is determined solely by its bias. 
The input-dependent routing is therefore inactive at initialization and is introduced gradually as the projection weights and the scalar gains $\alpha^{\mathrm{pre}}, \alpha^{\mathrm{post}}, \alpha^{\mathrm{res}}$ (which scale the dynamic term before the bias is added) are learned; the gains are initialized to a small value, $10^{-2}$.
The remaining biases make the block reduce, at initialization, to an ordinary residual sublayer $x_{l+1} = x_l + f_l(x_l)$.

\textbf{Write bias}: The write gate initializes $b^{\mathrm{post}}=0$, giving unit gates since $q = 2\sigma(0) = 1$. 
So, the sublayer (MLP/attention) output is written equally to all the streams. 

\textbf{Residual-mixing bias.} We set $b^{\mathrm{res}}$ to an identity-patterned logit (diagonal entries $1$, off-diagonal $0$). 
Sinkhorn normalization turns this into a doubly-stochastic matrix with diagonal $A_{ii} = e/(e{+}n{-}1)$ and off-diagonals $A_{ij} = 1/(e{+}n{-}1)$ ($A_{ii}\approx 0.73$ for $n{=}2$, and $0.48$ for $n{=}4$); the diagonal is larger than each off-diagonal by the constant factor $e$. 
These exact values are immaterial at initialization: because all $n$ streams hold identical copies of the pretrained hidden state, the residual mix $\sum_j A[i,j]\,x^{(j)}$ returns that shared value for \emph{any} row-stochastic $A$, which Sinkhorn guarantees. 
The identity-patterned bias therefore does not alter the function at initialization. 
A uniform $b^{\mathrm{res}}$ would likewise preserve the pretrained function.

\textbf{Read gate}: 
The read-gate bias $b^{\mathrm{pre}}$ is used to break the symmetry among streams. 
We use the same \textsc{uniform-noisy} scheme for \emph{both} the specialized and non-specialized models: each stream is initialized to read approximately equally, with target $\sigma(b^{\mathrm{pre}}_k) = 1/n$ at every layer, perturbed by small per-stream Gaussian noise on the logits ($\sigma_{\mathrm{noise}} = 10^{-2}$), after which the biases are renormalized so that $\sum_k \sigma(b^{\mathrm{pre}}_k) = 1$ exactly---keeping the stream aggregation a convex combination (and hence function-preserving) at initialization. 
 We deliberately apply the \emph{identical} initialization to both conditions so that the only difference between them is whether the retrieval heads read disjoint streams (specialized) or the pooled streams (non-specialized), removing initialization as a confound. 
  
All Sinkhorn normalizations use $20$ iterations and $\epsilon = 10^{-6}$. 
The exact values are reported in Table~\ref{tab:mhc-init}.

Finally, the mHC routing parameters are optimized with a \emph{separate, higher learning rate} than the rest of the network.
During fine-tuning we place all routing parameters---the three projections, their scalar gains, and the block's RMSNorm---in their own Adam parameter group at learning rate $10^{-3}$, while the ModernBERT backbone and the retrieval heads train at the base learning rate. 
This lets the small, randomly perturbed routing parameters adapt quickly enough to develop specialization within a single fine-tuning epoch without destabilizing the pretrained backbone.

\begin{table}[t]
    \centering
    \small
    \setlength{\tabcolsep}{4pt}
    \caption{Initialization of each mHC block. Projection weights are zero-initialized so routing starts static; the biases place the
    block near an identity residual layer. The read-gate bias uses the \textsc{uniform-noisy} scheme (target
    $\sigma(b^{\mathrm{pre}}_k)=1/n$ plus noise $\varepsilon\sim\mathcal{N}(0,\sigma_{\mathrm{noise}}^2)$, $\sigma_{\mathrm{noise}}{=}10^{-2}$, renormalized to sum to one) for both specialized and non-specialized
  models. At initialization this yields read gate $p \approx [1/n,\dots,1/n]$ (summing to $1$), write gate $q = [1,\dots,1]$, and a doubly-stochastic mixing matrix $A$ with its largest mass on the diagonal.}
    \label{tab:mhc-init}
    \begin{tabular}{lll}
    \toprule
    Parameter & Shape & Initialization \\
    \midrule
    $W^{\mathrm{pre}}$  & $n \times nd$   & $0$ \\
    $W^{\mathrm{post}}$ & $n \times nd$   & $0$ \\
    $W^{\mathrm{res}}$  & $n^2 \times nd$ & $0$ \\
    \midrule
    $b^{\mathrm{pre}}$  & $n$   & $\mathrm{logit}(1/n)+\varepsilon$, renorm. \\
    $b^{\mathrm{post}}$ & $n$   & $0$ \\
    $b^{\mathrm{res}}$  & $n^2$ & diag $1$, off-diag $0$ \\
    \midrule
    $\alpha^{\mathrm{pre/post/res}}$ & $1$ & $10^{-2}$ \\
    RMSNorm $\gamma$ & $nd$ & $1$ \\
    \bottomrule
    \end{tabular}
  \end{table}

\section{How specialization reshapes the streams (attention sublayer)}
\label{app:probe-attn}

\begin{figure*}[t]
    \centering
    \includegraphics[width=\textwidth]{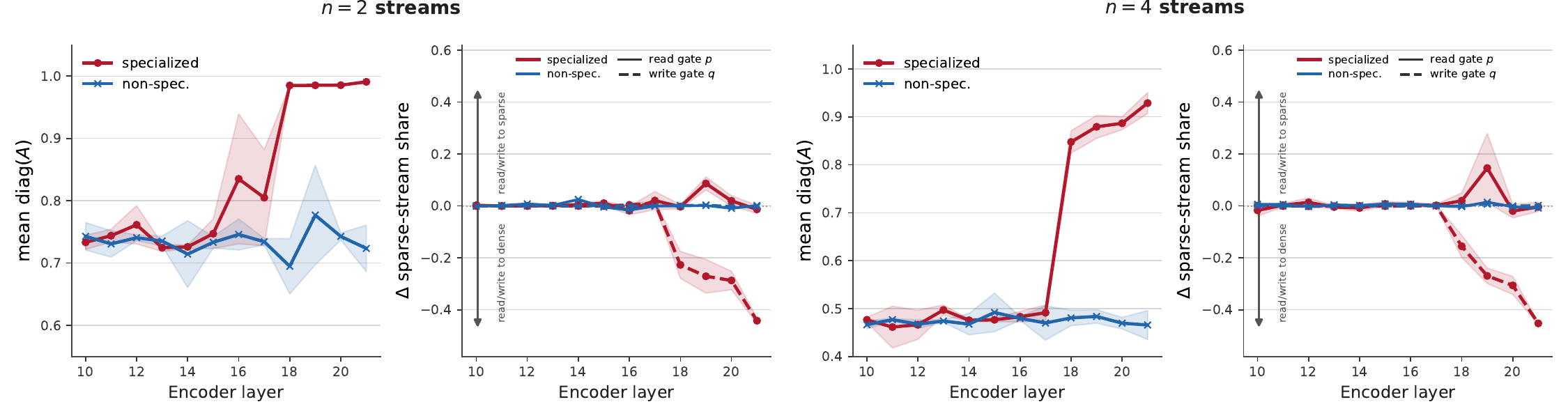}
    \caption{
    Attention-sublayer routing probes at the final checkpoint (lines are the mean and shaded bands $\pm 1$ standard deviation over 4 seeds; flat layers $0$--$9$ omitted). Left pair: $n{=}2$; right pair: $n{=}4$.
    Within each pair: mean diagonal of the residual-mixing matrix $\mathrm{diag}(A)$,
    and the shift in each gate's sparse-stream share $\Delta$ (read gate $p$ solid,
    write gate $q$ dashed). Red: specialized; blue: non-specialized control. Like the
    MLP sublayer, attention drives $A$ toward the identity under specialization; unlike
    the MLP, its read gate does not shift ($\Delta s(p_{\ell})\approx 0$) while its write gate
    moves strongly toward the dense-assigned stream(s) ($\Delta s(q_{L})\approx -0.44$),
    i.e.\ attention specializes its write but not its read. Metrics defined in
    Section~\ref{sec:depth}.
    }
    \label{fig:dynamics-row-mhc4}
\end{figure*}

Section~\ref{sec:depth} reports the routing probes for the MLP sublayer; here we provide the corresponding residual-mixing and read/write gate probes for the attention sublayer.

Figure~\ref{fig:dynamics-row-mhc4} shows that attention mirrors the MLP in the residual-mixing gate: under specialization, $A$ moves toward the identity in the top layers (mean diagonal $\approx 0.99$ at $n{=}2$ and $\approx 0.93$ at $n{=}4$, versus $\approx 0.72$ and $\approx 0.47$ for the non-specialized control). 
The gate behavior, however, differs from the MLP counterpart. 
Whereas the MLP shifts its read gate toward the sparse stream and its write gate toward the dense stream, attention leaves its read gate essentially unchanged ($\Delta s(p_{\ell})\approx 0$) and moves only its write gate---strongly---toward the dense-assigned stream(s) ($\Delta s(q_{L})\approx -0.44$ at $n{=}2$ and $-0.45$ at $n{=}4$). 
Attention therefore specializes by routing its \emph{write} rather than its read: it keeps reading from all streams but commits its output predominantly to the dense streams. 

As in the main text, these patterns are consistent across both stream counts and all four seeds.
\section{Results at a Larger Model Scale}
\label{app:large}

Table~\ref{tab:beir-large} reports the same comparison as Table~\ref{tab:beir-categories}, using ModernBERT-large in place of ModernBERT-base. The specialization advantage reported in Section~\ref{sec:mo-retrieval} holds at this larger scale as well.

\begin{table*}[t]
\centering
\small
\resizebox{\textwidth}{!}{%
\begin{tabular}{lcccccccccccccc}
\toprule
Model & ArguAna & C-FEVER & DBPedia & FEVER & FiQA & HotpotQA & NFC & NQ & Quora & SciDocs & SciF & TREC-C & Touch\'e & Avg \\
\midrule
\multicolumn{15}{l}{$n=1$} \\
Sparse-only  & 0.3967 & 0.0954 & 0.1348 & 0.5931 & 0.2133 & 0.5568 & 0.2251 & 0.1654 & 0.6761 & 0.1118 & 0.6507 & 0.5178 & 0.2307 & 0.3514 \\
Dense-only   & 0.4110 & 0.2157 & \underline{0.3072} & 0.5351 & \underline{0.3103} & 0.4061 & 0.2800 & \underline{0.5063} & \underline{0.8562} & \underline{0.1611} & 0.5972 & 0.6996 & 0.2205 & 0.4236 \\
Sparse+Dense & \underline{0.4268} & \underline{0.2204} & 0.2711 & \underline{0.6586} & 0.3095 & \underline{0.5838} & \underline{0.2828} & 0.4488 & 0.8443 & 0.1398 & \textbf{\underline{0.6970}} & \underline{0.7614} & \underline{0.2939} & \underline{0.4568} \\
\midrule
\multicolumn{15}{l}{$n=2$} \\
mHC no-spec & 0.4103 & 0.2108 & 0.2838 & \textbf{\underline{0.6698}} & 0.3076 & \underline{0.5864} & 0.2983 & 0.4546 & 0.8453 & 0.1421 & \underline{0.6911} & \underline{0.7621} & \underline{0.2836} & 0.4574 \\
mHC spec    & \underline{0.4509} & \underline{0.2285} & \underline{0.3237} & 0.6353 & \underline{0.3448} & 0.5332 & \textbf{\underline{0.3115}} & \underline{0.5182} & \underline{0.8627} & \textbf{\underline{0.1655}} & 0.6889 & 0.7534 & 0.2506 & \underline{0.4667} \\
\midrule
\multicolumn{15}{l}{$n=4$} \\
mHC no-spec & 0.4027 & 0.2001 & 0.2814 & \underline{0.6669} & 0.3081 & \textbf{\underline{0.5887}} & 0.2984 & 0.4573 & 0.8437 & 0.1358 & \underline{0.6940} & 0.7721 & \textbf{\underline{0.2954}} & 0.4573 \\
mHC spec    & \textbf{\underline{0.4832}} & \textbf{\underline{0.2389}} & \textbf{\underline{0.3270}} & 0.6438 & \textbf{\underline{0.3544}} & 0.5447 & \underline{0.3101} & \textbf{\underline{0.5200}} & \textbf{\underline{0.8637}} & \underline{0.1628} & 0.6901 & \textbf{\underline{0.7802}} & 0.2531 & \textbf{\underline{0.4748}} \\
\bottomrule
\end{tabular}%
}
\caption{BEIR nDCG@10 across all 13 datasets using ModernBERT-large, mean over two seeds (42 and 1234). Sparse-only and Dense-only are evaluated in their native retrieval mode; Sparse+Dense and the mHC models are hybrid retrievers fused at $\alpha=0.9$. \underline{Underline} marks the best score within each stream count $n$; \textbf{bold} marks the best score for that dataset across all configurations.}
\label{tab:beir-large}
\end{table*}
\section{Per-Seed Results}
\label{app:seeds}

Table~\ref{app:seeds} and Table~\ref{tab:per-seed-large} reports BEIR nDCG@10 for each of the two training seeds separately, across all 13 datasets at
fusion weight $\alpha=0.9$; the two-seed means are the values reported in Table~\ref{tab:beir-categories} and Table~\ref{tab:beir-large} for ModernBERT-base and ModernBERT-large respectively. 
The specialization gain is consistent across seeds: at both $n=2$ and $n=4$ and at both scales: ModernBERT-base and ModernBERT-large.

\begin{table*}[t]
  \centering
  \small
  \resizebox{\textwidth}{!}{%
  \begin{tabular}{llcccccccccccccc}
  \toprule
  Model & Seed & ArguAna & C-FEVER & DBPedia & FEVER & FiQA & HotpotQA & NFC & NQ & Quora & SciDocs & SciF & TREC-C & Touch\'e & Avg
  \\
  \midrule
  \multicolumn{16}{l}{$n=1$} \\
  Sparse-only  & 42   & 0.4038 & 0.1040 & 0.1504 & 0.6151 & 0.1998 & 0.5857 & 0.2238 & 0.1637 & 0.6823 & 0.1134 & 0.6555 & 0.5704 & 0.2269 & 0.3611 \\
               & 1234 & 0.3849 & 0.1064 & 0.1463 & 0.6617 & 0.1908 & 0.5932 & 0.2200 & 0.1562 & 0.6620 & 0.1105 & 0.6428 & 0.5057 & 0.2476 & 0.3560 \\[5pt]
  Dense-only   & 42   & 0.3946 & 0.2119 & 0.2690 & 0.5025 & 0.2419 & 0.3785 & 0.2529 & 0.4606 & 0.8488 & 0.1315 & 0.5600 & 0.6725 & 0.2214 & 0.3959 \\
               & 1001 & 0.4003 & 0.2112 & 0.2715 & 0.5072 & 0.2326 & 0.3808 & 0.2392 & 0.4532 & 0.8471 & 0.1341 & 0.5426 & 0.6843 & 0.2187 & 0.3941 \\[5pt]
  Sparse+Dense & 42   & 0.3944 & 0.2022 & 0.2581 & 0.6382 & 0.2676 & 0.5657 & 0.2622 & 0.4047 & 0.8289 & 0.1309 & 0.6622 & 0.6797 & 0.2615 & 0.4274 \\
               & 1234 & 0.3723 & 0.2412 & 0.2523 & 0.6669 & 0.2622 & 0.5700 & 0.2601 & 0.4173 & 0.8258 & 0.1311 & 0.6793 & 0.7184 & 0.2671 & 0.4357 \\
  \midrule
  \multicolumn{16}{l}{$n=2$} \\
  mHC no-spec & 42   & 0.3834 & 0.2054 & 0.2605 & 0.6465 & 0.2744 & 0.5744 & 0.2697 & 0.4127 & 0.8253 & 0.1352 & 0.6731 & 0.7084 &
  0.2576 & 0.4328 \\
              & 1234 & 0.3819 & 0.2211 & 0.2553 & 0.6593 & 0.2693 & 0.5648 & 0.2719 & 0.4030 & 0.8205 & 0.1339 & 0.6688 & 0.6601 &
  0.2692 & 0.4292 \\[5pt]
  mHC spec    & 42   & 0.4130 & 0.2329 & 0.2907 & 0.6248 & 0.2883 & 0.5243 & 0.2801 & 0.4676 & 0.8472 & 0.1510 & 0.6755 & 0.7726 &
  0.2513 & 0.4476 \\
              & 1234 & 0.4427 & 0.2459 & 0.2931 & 0.6340 & 0.2887 & 0.5209 & 0.2732 & 0.4685 & 0.8494 & 0.1560 & 0.6589 & 0.7444 &
  0.2682 & 0.4495 \\
  \midrule
  \multicolumn{16}{l}{$n=4$} \\
  mHC no-spec & 42   & 0.4183 & 0.2237 & 0.2586 & 0.6537 & 0.2738 & 0.5733 & 0.2677 & 0.4196 & 0.8294 & 0.1384 & 0.6743 & 0.7772 &
  0.2548 & 0.4433 \\
              & 1234 & 0.3954 & 0.2371 & 0.2582 & 0.6805 & 0.2766 & 0.5698 & 0.2712 & 0.4112 & 0.8308 & 0.1348 & 0.6759 & 0.6789 &
  0.2827 & 0.4387 \\[5pt]
  mHC spec    & 42   & 0.4411 & 0.2356 & 0.2980 & 0.6270 & 0.2868 & 0.5210 & 0.2821 & 0.4726 & 0.8511 & 0.1549 & 0.6591 & 0.7765 &
  0.2568 & 0.4510 \\
              & 1234 & 0.4164 & 0.2482 & 0.3002 & 0.6501 & 0.2876 & 0.5242 & 0.2874 & 0.4788 & 0.8520 & 0.1490 & 0.6787 & 0.7619 &
  0.2785 & 0.4548 \\
  \bottomrule
  \end{tabular}%
  }
  \caption{Per-seed BEIR nDCG@10 across all 13 datasets (hybrid fusion at $\alpha=0.9$) for ModernBERT-base.}
  \label{app:seeds}
  \end{table*}

\begin{table*}[t]
\centering
\small
\resizebox{\textwidth}{!}{%
\begin{tabular}{llcccccccccccccc}
\toprule
Model & Seed & ArguAna & C-FEVER & DBPedia & FEVER & FiQA & HotpotQA & NFC & NQ & Quora & SciDocs & SciF & TREC-C & Touch\'e & Avg \\
\midrule
\multicolumn{16}{l}{$n=1$} \\
Sparse-only  & 42   & 0.4111 & 0.0981 & 0.1326 & 0.5833 & 0.2124 & 0.5491 & 0.2278 & 0.1701 & 0.6799 & 0.1143 & 0.6582 & 0.5424 & 0.2262 & 0.3543 \\
             & 1234 & 0.3822 & 0.0926 & 0.1371 & 0.6028 & 0.2142 & 0.5644 & 0.2223 & 0.1608 & 0.6722 & 0.1092 & 0.6433 & 0.4932 & 0.2351 & 0.3484 \\
Dense-only   & 42   & 0.4107 & 0.2158 & 0.3080 & 0.5403 & 0.3078 & 0.4052 & 0.2777 & 0.5047 & 0.8563 & 0.1606 & 0.5944 & 0.7053 & 0.2233 & 0.4239 \\
             & 1234 & 0.4112 & 0.2156 & 0.3063 & 0.5299 & 0.3127 & 0.4070 & 0.2823 & 0.5079 & 0.8562 & 0.1616 & 0.5999 & 0.6938 & 0.2178 & 0.4232 \\
Sparse+Dense & 42   & 0.4291 & 0.2133 & 0.2670 & 0.6538 & 0.3027 & 0.5829 & 0.2872 & 0.4477 & 0.8459 & 0.1391 & 0.7031 & 0.7472 & 0.2915 & 0.4547 \\
             & 1234 & 0.4245 & 0.2275 & 0.2751 & 0.6634 & 0.3163 & 0.5848 & 0.2783 & 0.4500 & 0.8427 & 0.1405 & 0.6909 & 0.7756 & 0.2963 & 0.4589 \\
\midrule
\multicolumn{16}{l}{$n=2$} \\
mHC no-spec & 42   & 0.3971 & 0.2038 & 0.2833 & 0.6553 & 0.3044 & 0.5937 & 0.2932 & 0.4537 & 0.8459 & 0.1389 & 0.6843 & 0.7649 & 0.2830 & 0.4540 \\
            & 1234 & 0.4234 & 0.2178 & 0.2842 & 0.6842 & 0.3109 & 0.5791 & 0.3033 & 0.4554 & 0.8447 & 0.1454 & 0.6979 & 0.7593 & 0.2843 & 0.4608 \\
mHC spec    & 42   & 0.4417 & 0.2375 & 0.3270 & 0.6416 & 0.3464 & 0.5386 & 0.3099 & 0.5229 & 0.8626 & 0.1646 & 0.6845 & 0.7676 & 0.2593 & 0.4696 \\
            & 1234 & 0.4602 & 0.2195 & 0.3205 & 0.6290 & 0.3432 & 0.5277 & 0.3131 & 0.5135 & 0.8627 & 0.1664 & 0.6934 & 0.7393 & 0.2420 & 0.4639 \\
\midrule
\multicolumn{16}{l}{$n=4$} \\
mHC no-spec & 42   & 0.4103 & 0.2011 & 0.2815 & 0.6683 & 0.2958 & 0.5877 & 0.2997 & 0.4519 & 0.8412 & 0.1358 & 0.6894 & 0.7717 & 0.2880 & 0.4556 \\
            & 1234 & 0.3950 & 0.1992 & 0.2813 & 0.6655 & 0.3204 & 0.5898 & 0.2970 & 0.4627 & 0.8462 & 0.1359 & 0.6986 & 0.7725 & 0.3029 & 0.4590 \\
mHC spec    & 42   & 0.4805 & 0.2372 & 0.3268 & 0.6467 & 0.3526 & 0.5494 & 0.3079 & 0.5125 & 0.8621 & 0.1632 & 0.6870 & 0.7747 & 0.2526 & 0.4733 \\
            & 1234 & 0.4859 & 0.2406 & 0.3273 & 0.6408 & 0.3562 & 0.5400 & 0.3123 & 0.5276 & 0.8652 & 0.1625 & 0.6932 & 0.7856 & 0.2536 & 0.4762 \\
\bottomrule
\end{tabular}%
}
\caption{Per-seed BEIR nDCG@10 across all 13 datasets using ModernBERT-large, for each model and training seed. Sparse-only and Dense-only are evaluated in their native retrieval mode; the remaining models are hybrid retrievers fused at $\alpha=0.9$. The two-seed means are the values reported in Table~\ref{tab:beir-large}. As at the base scale, the specialization gain holds within each seed: at both $n=2$ and $n=4$, mHC+spec exceeds mHC no-spec in every seed.}
\label{tab:per-seed-large}
\end{table*}

\end{document}